\documentclass{article} 
\usepackage{iclr2017_workshop,times}
\usepackage{hyperref}
\usepackage{url}
\usepackage{caption}
\usepackage{subcaption}
\usepackage{graphicx}

\usepackage{hyperref}
\usepackage[english]{babel}
\usepackage{url}
\usepackage{mathtools}
\usepackage{amsmath}
\usepackage{amsthm}
\usepackage{amssymb}
\usepackage{wrapfig}
\usepackage{bm}
\usepackage{psfrag}
\usepackage{tikz}
\usepackage{epstopdf}
\usepackage{adjustbox}
\usepackage{multirow}

\newcommand{\beq}[1][\vspace{0.3em}]{#1\begin{equation}}
\newcommand{\eeq}{\end{equation}}

\newcommand{\bit}{\vspace{0mm}\begin{itemize}}
\newcommand{\eit}{\vspace{0mm}\end{itemize}}
\newcommand{\ben}{\vspace{0mm}\begin{enumerate}}
\newcommand{\een}{\vspace{0mm}\end{enumerate}}

\newcommand{\Xv}[0]{{{\bf X}}}
\newcommand{\Yv}[0]{{{\bf Y}}}

\newcommand{\xv}[0]{{{\bf x}}}
\newcommand{\yv}[0]{{{\bf y}}}
\newcommand{\zv}[0]{{{\bf z}}}

\newcommand{\bb}[1]{\mathbb{#1}}

\newcommand{\mc}[1]{\mathcal{#1}}

\title{
On the Limits of Learning Representations with Label-Based Supervision
}

\author{Jiaming Song, Russell Stewart, Shengjia Zhao \& Stefano Ermon \\
Computer Science Department\\
Stanford University\\
\texttt{\{tsong,stewartr,zhaosj12,ermon\}@cs.stanford.edu} \\
}

\begin{document}
\maketitle
\section{Introduction}
Advances in neural network based classifiers have transformed automatic feature learning from a pipe dream of stronger AI to a routine and expected property of practical systems. Since the emergence of AlexNet \citep{krizhevsky2012imagenet}, every winning submission of the ImageNet challenge \citep{russakovsky2015imagenet} has employed end-to-end representation learning, and due to the utility of good representations for transfer learning \citep{yosinski2014transferable}, representation learning has become as an important and distinct task from supervised learning. At present, this distinction is inconsequential, as supervised methods are state-of-the-art in learning transferable representations~\citep{nguyen2016plug,salimans2016improved}. But recent work \citep{radford2015unsupervised,chen2016infogan} has shown that generative models can also be powerful agents of representation learning. Will the representations learned from these generative methods ever rival the quality of those from their supervised competitors? In this work, we argue in the affirmative, that from an information theoretic perspective, generative models have greater potential for representation learning. Based on several experimentally validated assumptions, we show that supervised learning is upper bounded in its capacity for representation learning in ways that certain generative models, such as Generative Adversarial Networks (GANs \citet{goodfellow2014generative}) are not. We hope that our analysis will provide a rigorous motivation for further exploration of generative representation learning. 

\section{Feature Learning with Discriminative Models}
\label{sec:supervised}

Let $\xv \in \bb{R}^d$ be observations drawn from a distribution $p_{\Xv}(\xv)$, and $\yv \in \bb{R}^\ell$ be labels for $\xv$ obtained through a deterministic mapping $\yv = g(\xv)$.
Assume we are operating in some domain (e.g. computer vision), and there exists a set of good features $\mc{F}_g$ that we would like to learn (e.g. a feature that denotes ``tables''). These features will emerge from suitable weights of a deep neural network (e.g. a filter that detects tables), and thus must compete against an exponentially large set of bad, random features. 
Our goal is to learn all the good features from the dataset in the process of using a neural network to perform certain tasks.

We analyze the feature learning process by parameterizing the state of a network according to the set of features it has already learned. We then investigate the marginal value of learning an additional feature. If we have thus far learned $k-1$ features, $\{f_i\}_{i=1}^{k-1}$, we propose to measure the ease of learning the $k$-th feature according to the reduction in entropy of the labels when we add the new feature to improve the supervised learning performance.
\begin{equation}
\text{signal}(f_k) = I(\Yv; f_k(\Xv) | f_1(\Xv), \ldots, f_{k-1}(\Xv)) \footnote{In the remainder of the paper, We remove the $\Xv$ in $f(\Xv)$ to ease notation.} \label{eq:signal}
\end{equation}

This concept simply encodes our intuition that features will be easier to learn when they pertain more directly to the task at hand, and it aligns well with the ``information gain'' feature selection metric in Random Forests and Genomic studies \citep{schleper2005genomic}.


If we are willing to believe that the learnability of the feature corresponds to its signal (as we argue below), we observe the following upper bound on the potential for learning features using labeled supervision. If we aim to learn $k$ features, then the sum of signal over all of those features must be no greater than the entropy in the labels, since
\begin{equation}
\sum_{i=1}^{k} \text{signal}(f_i) = I(\Yv; f_1, \ldots, f_k) \leq H(\Yv) \label{eq:signal-bound}
\end{equation}
But then at least some of those features must be very hard to learn, as $\min_{i \in \{1, \ldots, k\}} \text{signal}(f_i) \leq H(\Yv) / k$.
Thus, independent of the size of our dataset, we have an upper bound on the capacity of our model to learn a large number of features. In the remainder of the paper, we will refer to this phenomenon as ``feature competition''.



\subsection{Experimental Validation of Feature Competition}
We tested the ``feature competition'' hypothesis with the following experiment on MNIST digits. Each input image $\xv$ contained two side by side digits randomly selected from the MNIST dataset, denoted as $\xv_{l}$ (left) and $\xv_{r}$ (right). We denote their respective ground truth labels as $\yv_{l}$ and $\yv_{r}$. 

We first trained a feature detector $f$ with inputs $\xv = (\xv_l, \xv_r)$, using only the {\em left} label $\yv_l$ as supervision. Then given $f$, we train a logistic regression classifier to predict the {\em right} label $\yv_r$ with inputs $\zv = f(\xv)$, keeping $f$ fixed. If $\yv_l$ and $\yv_r$ are uncorrelated, this task is very difficult, as there no value to learning features from $\xv_r$ in the original task. But here, we show that even if features from $\xv_r$ contain information about $\yv_l$, sufficiency of features from $\xv_l$ decreases the conditional mutual information of features from $\xv_r$ and makes them harder to learn.


We consider $f(\xv) = [f_1(\xv_l); f_2(\xv_r)]$ to be an MLP, with $f_1$ and $f_2$ being two separate networks.
$f$ is the concatenation of $f_1$ and $f_2$, which is used to perform supervised learning tasks. In the first phase, we 1) completely corrupt \footnote{The corruption is done by sampling $\xv_l$ from a factored Gaussian distribution, where the mean and variance corresponds to the mean and variance of the MNIST training set.} the left digit $\xv_l$ with some probability $(1 - \rho_l)$, and 2) assign the right digit and label to have the same label as the left digit with probability $\rho_r$. 
In the pretraining task we are able to calculate the signal for learning features in $\xv_r$, which is $I(\Yv_l; \Xv_r | \Xv_l)$,
 a function of $\rho_l$ and $\rho_r$. 



\begin{figure}
\centering
\begin{subfigure}[t]{0.33\textwidth}
\centering
\includegraphics[width=\textwidth]{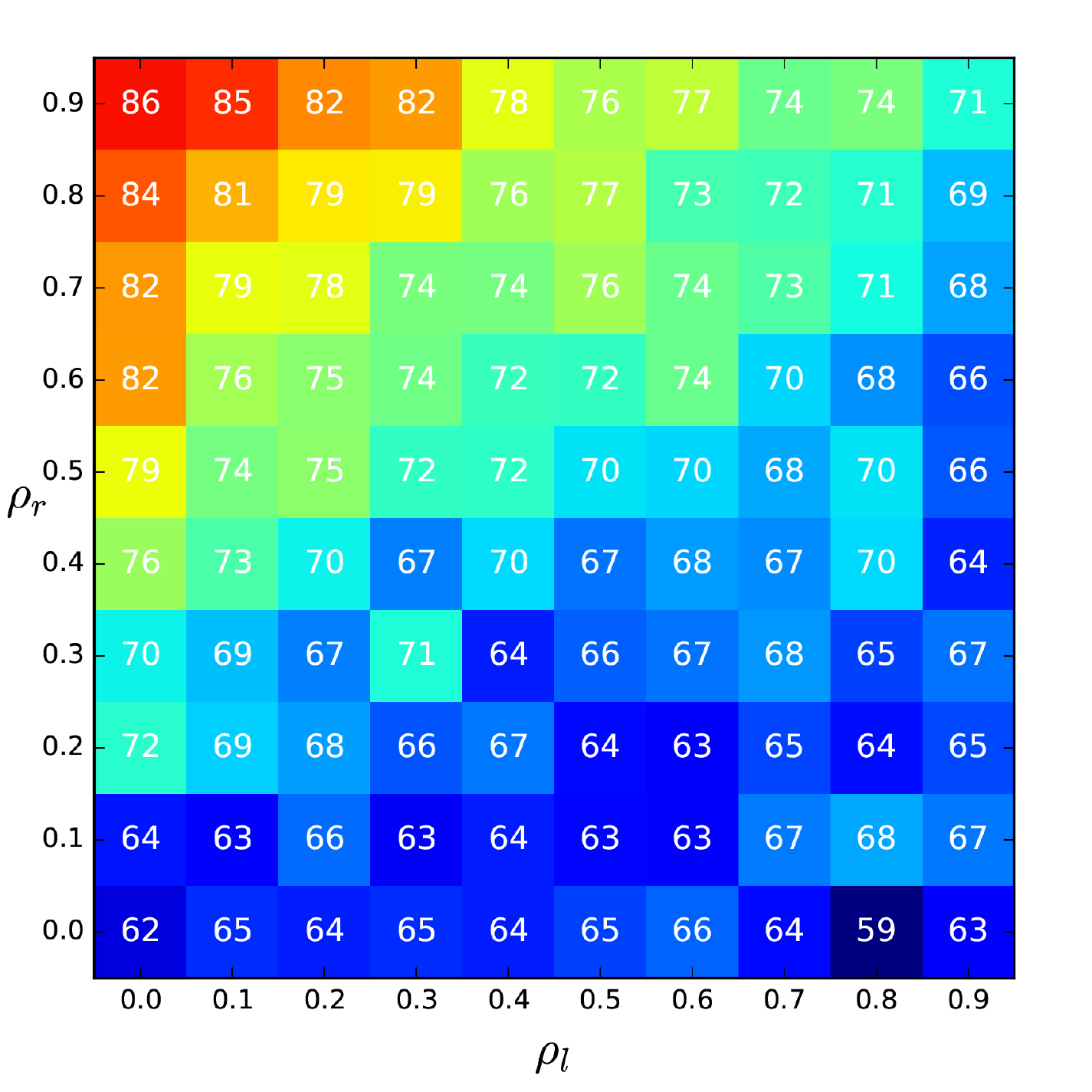}
\caption{Test accuracy over $\yv_r$ with different $\rho_l$ and $\rho_r$.}
\end{subfigure}
~
\begin{subfigure}[t]{0.33\textwidth}
\centering
\includegraphics[width=\textwidth]{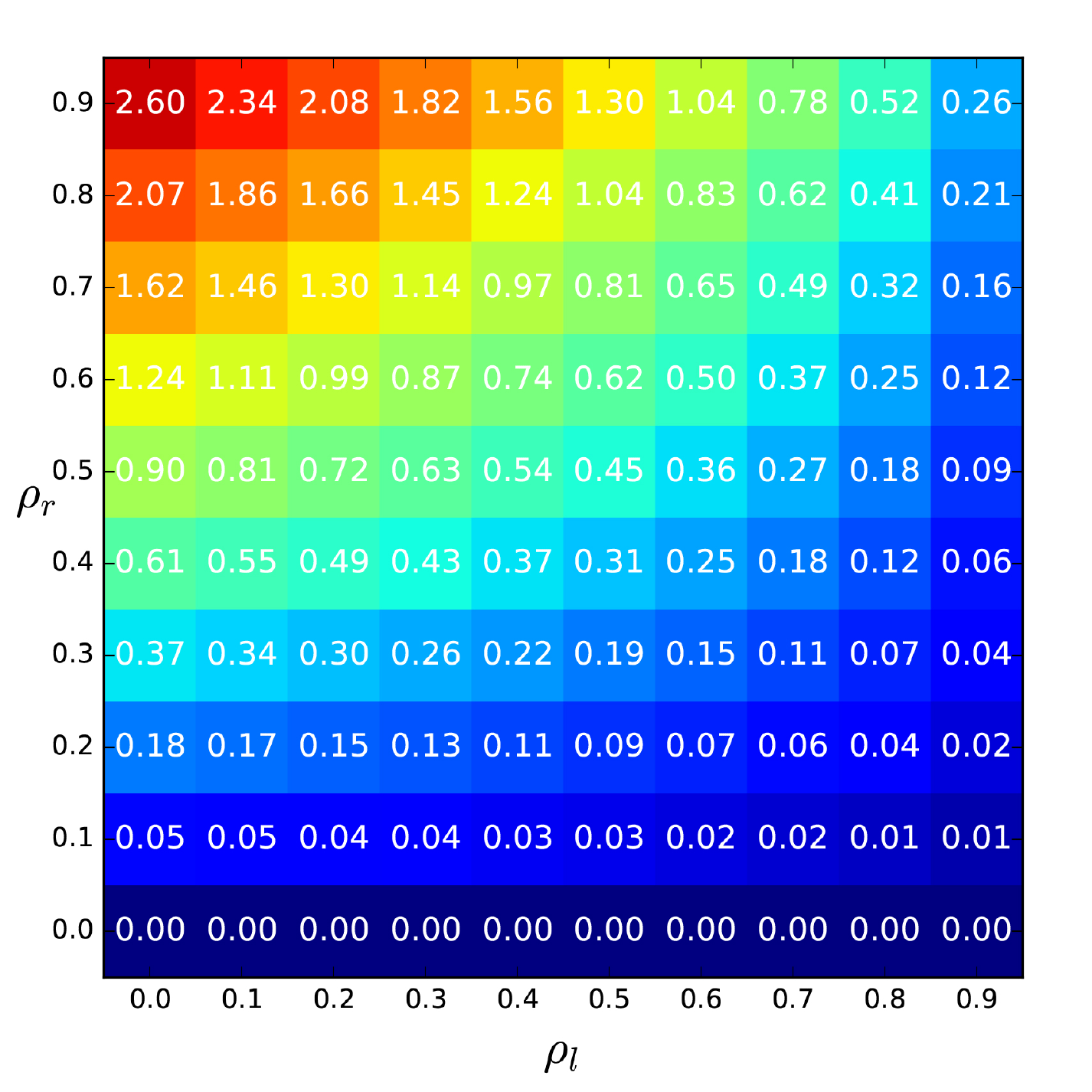}
\caption{Calculated $I(\Yv_l ; \Xv_r| \Xv_l)$ given $\rho_l$ and $\rho_r$.}
\end{subfigure}
\caption{Test accuracy over $\yv_r$ and $I(\Yv; \Xv_r| \Xv_l)$. The r-value between test accuracy and signals is 0.9213, which suggests that features having higher conditional mutual information with the labels are easier to learn.}
\label{fig:rho}
\end{figure}

Given a fixed $\rho_r$, increasing $\rho_l$ will make the features for $\xv_l$ easier to learn, without changing the relationship between $\xv_r$ and $\yv_l$. However, as shown in Figure \ref{fig:rho} for any fixed $\rho_r$, increasing $\rho_l$ will decrease the test performance in the second phase. This suggests that the ability to learn high quality features from $\xv_l$ decreases the ability to learn features from $\xv_r$, and thus there is a direct competition between these two features in supervised learning.

\section{Feature Learning with Generative Models}
If supervised learning is bounded in its capacity for feature extraction by the entropy of the labels, what of unsupervised learning? In this section, we show that one family of unsupervised learning methods, Generative Adversarial Networks (GAN \citet{goodfellow2014generative}), is not impacted by feature competition under limited assumptions.

In GANs, a generator network, $G$, generates samples (with generative distribution $\mc{G}$), and a discriminator network, $D(\xv)$, attempts to distinguish between those samples and real data. 
$D$ is presented with a labeled dataset $\{(\xv_i, y_i)\}_{i=1}^{m}$, where $y = 1$ if $\xv \sim \mc{D}$, and $y = 0$ if $\xv \sim \mc{G}$; the two classes are balanced, so $H(y) = 1$. \footnote{In this section, we consider the distribution over $\xv$ to be the average of $\mc{G}$ and $\mc{D}$ for entropy and mutual information terms.}

Assume that $G$ and $D$ have already learned $k-1$ features $f_1, \ldots, f_{k-1}$, where the discriminator cannot separate samples from $\mc{G}$ and $\mc{D}$ with only these features. This indicates that $\Pr(y = 1|f_1(\xv), \cdots, f_{k-1}(\xv)) = 0.5$ for all $\xv$ in the dataset,
and that $H(y|f_1, \ldots, f_{k-1}) = 1$. Thus, the discriminator is in a state of confusion. We measure the motivation of $D$ for learning a new feature $f_k$ to be 
\begin{align}
I(y; f_k | f_1, \ldots, f_{k-1}) &= H(y | f_1, \ldots, f_{k-1}) - H(y | f_1, \ldots, f_{k-1}, f_k) \nonumber \\
 & = 1 - H(y|f_1, \ldots, f_{k-1}, f_k) \nonumber \\
 & \geq 1 - H(y | f_k) \label{eq:sig-disc}
\end{align}

where $H(y | f_k) \in [0, 1]$ is a measure of similarity between the distributions on the feature $f_k$ for real and generated samples. $H(y | f_k) = 1$ if and only if $f_k(\xv)$ is identically distributed almost everywhere for $G$ and $D$. Thus, if the generated distribution does not yet match the real data distribution along $f_k$, we will have positive signal to learn. Importantly, this lower bound has no dependence on previously learned features, $f_1, \ldots, f_{k-1}$. That is, when the discriminator is in a state of confusion, we have no feature competition.

\begin{table}
\centering
\begin{tabular}{|c|c|c|c|c|c|}
\hline
Model & CNN & AE & GAN & WGAN \\\hline
\multirow{2}{*}{Accuracy} & 67.93 & 89.95 & 90.38 & 91.37 \\
& (84.31) & (82.18) & (82.27) & (84.97) \\\hline
\end{tabular}
\caption{Test accuracy for $\yv_r$, given $f$ learned by different architectures. The numbers in the brackets indicate using weights that are not trained, which is a baseline for the case of random features.}
\label{table:unsupervised}
\end{table}

\subsection{Experimental Validation of Generative Models}
We have shown theoretically that one class of generative models, GANs, is not limited by the same upper bound on feature learning signal as discriminative models such as CNNs. We now empirically test these implications by revisiting the two-digit experiment from Section \ref{sec:supervised}. We set $\rho_l = 1$ and $\rho_r = 0$, where the two digits are selected completely at random (and where feature learning for $\xv_r$ using supervised learning methods performed worst). We consider four frameworks - a feed forward convolutional net (CNN); a traditional GAN \citep{goodfellow2014generative}; a recently proposed Wasserstein GAN (WGAN, \citet{arjovsky2017wasserstein}) and an Autoencoder\footnote{We do not split $f$ into two networks in the convolution setting.}. For the four frameworks, we use the same CNN architecture and set the output of $f$ to be the 100 neurons at the second top layer.
The results, shown in Table ~\ref{table:unsupervised}, demonstrate that in spite of the absence of labels, the features learned by all three generative models we considered, including GANs, AEs, and WGANs, were useful in the subsequent task of learning to recognize the right digit.

\section{Conclusion}
We have identified an upper bound on the capacity of supervised models to extract features which depends not on the size of the dataset, but rather the quality of labels. Our results suggest great promise for the future of feature extraction with unlabeled approaches.


\bibliography{iclr2017_workshop}
\bibliographystyle{iclr2017_workshop}
\newpage
\appendix

\section{Motivation for Balancing GANs via Loss Statistics}
\label{sec:motivation}
Assume that the ``state of confusion'' assumption breaks for $D$, such that $D$ has learned $l > 1$ more features than $G$ has learned, and it can classify better than random guessing. Therefore, $H(y|f_1, \ldots, f_{k-1}) = 1$, $H(y|f_1, \ldots, f_{k+l-1}) < 1$, and
\begin{equation}
H(y|f_k, \ldots, f_{k+l-1}) < 1 \label{eq:confusion-break}
\end{equation}
The motivation of $D$ for learning a new feature $f_{k+l}$ then becomes
\begin{align}
I(y; f_{k+l} | f_1, \ldots, f_{k+1-1}) &=\ H(y|f_1, \ldots, f_{k+l-1}) - H(y|f_1, \ldots, f_{k+l}) \nonumber \\
&=\ H(y|f_k, \ldots, f_{k+l-1}) - H(y|f_k, \ldots, f_{k+l})
\end{align}
which is no longer independent of $f_k, \ldots, f_{k+l-1}$ because of Equation \ref{eq:confusion-break}. This is analogous to the supervised learning setting - $D$ is simply trying to learn a new features $f_{k+l}$, given all the previous features $f_{k}, \ldots, f_{k+l-1}$ to optimize a fixed objective defined by features $f_1, \ldots, f_{k-1}$.

If $D$ has learned $k+l - 1$ features and $G$ has learned $k-1$ features, then $G$ is motivated to learn the proper distribution for feature $f_k$ to minimize $H(y|f_1, \ldots, f_{k+l-1})$. However, this quantity will still be smaller than one even if we assume $G$ learns the correct distribution on $f_k$, so the incentive for $G$ becomes \footnote{We use $H_{f_k}$ to denote the expectation is performed over the distribution when $G$ has learned the proper distribution for feature $f_k$.}
\begin{align}
&\ H_{f_k}(y|f_1, \ldots, f_{k+l-1}) - H(y|f_1, \ldots, f_{k+l-1}) \nonumber \\
< &\ 1 - H(y|f_1, \ldots, f_{k+l-1}) \nonumber \\
= &\ H(y) - H(y|f_{k}, \ldots, f_{k+l-1}) = I(y; f_{k}, \ldots, f_{k+l-1})
\end{align}

Notice that the mutual information $I(y; f_{k}, \ldots, f_{k+l-1})$ is exactly the sum of motivation for the discriminator to learn features $f_k, \ldots, f_{k+l}$. This implies that if we continue to allow $D$ and $G$ to learn one feature at a time, which is the case where we do not attempt to balance GANs via loss statistics, $G$ will not catch up with $D$ in one step; $D$, on the other hand, the advantage in $D$ will cause it to suffer from the ``feature competition'' challenge, where it has less incentive to learn features than it should.

One obvious method to counter this is by balancing GANs via loss statistics; although $D$ and $G$ suffers from feature competition during learning $f_k, \ldots, f_{k+l-1}$, $G$ will catch up if it learns multiple features consecutively, so that it makes $D$ confused again, where $V(D, G) = \log 4$. However, we are not promoting the strategy where $G$ and $D$ should be trained more whenever its loss exceeds some predetermined value, but we believe principled approaches to tackle this problem will be valuable to training of GANs.

\end{document}